\documentclass[pmlr]{jmlr}% new name PMLR (Proceedings of Machine Learning)

\usepackage{longtable}% for long tables

\usepackage{booktabs}
\usepackage{tikz}
\usetikzlibrary{graphs,arrows,positioning,calc,trees,decorations.pathmorphing}
\tikzstyle{hidden} = [circle,draw]
\tikzstyle{visible} = [circle,draw,fill=black!10]
\tikzstyle{myLink} = [->,>=latex]
\tikzstyle{myLink2} = [<->,>=latex]

\jmlrvolume{1}
\jmlryear{2018}
\jmlrworkshop{Learning and Automata (LearnAut) 2018}

\title[Learning Tree Distributions by HMMs]{Learning Tree Distributions by Hidden Markov Models}

  \author{\Name{Davide Bacciu} \Email{bacciu@di.unipi.it} \\
   \Name{Daniele Castellana} \Email{daniele.castellana@di.unipi.it}\\
   \addr Dipartimento di Informatica - Universit{\`a} di Pisa}

\editor{Editor's name}
\begin{document}

\maketitle

\begin{abstract}
Hidden tree Markov models allow learning distributions for tree structured data while being interpretable as nondeterministic automata. We provide a concise summary of the main approaches in literature, focusing in particular on the causality assumptions introduced by the choice of a specific tree visit direction. We will then sketch a novel non-parametric generalization of the bottom-up hidden tree Markov model with its interpretation as a nondeterministic tree automaton with infinite states.
\end{abstract}

\section{Introduction}
\label{sec:intro}
Hidden Markov Models (HMMs) are popular generative models for sequential data, that is the simplest form of structured data characterized by the existence of a total ordering relation (e.g. often time) between the atomic elements of the sequence. As such HMMs allow to learn distributions on a space of sequential-structured samples. On the other hand, they have also a well studied characterization as string automata. \cite{dupont} characterized the links between classical HMMs and Probabilistic Automata (PA), showing that HMMs are equivalent to nondeterministic PA with finite states (PNFA). In particular, it can be shown that Markov models can be transformed into an equivalent PNFA with the same number of states and that this implication can be reversed provided that we relax the constraint on maintaining the same number of states.

Markov models have been generalized to learn distributions on trees, that are a form of structured data useful to naturally represent atomic entities bound by partial order relationships (e.g. hierarchy, containment, etc.), such as parse trees in natural language, abstract syntax trees, phylogenetic trees in biology, etc. When moving from the sequential to the tree-structured domain, the direction of the parsing process (in case of automata) or of the generative process (in case of probabilistic models) becomes relevant to characterize the capabilities of the computational model. Sequences can be processed both left-to-right and right-to-left, without any consequence on the representation capabilities of the computational model. Trees can be processed either top-down, i.e. from the root to the leaves, or bottom-up, i.e. from the leaves to the root: choosing one direction over the other has articulated consequences depending on the computational model used to process the tree.  On the one hand, the tree language recognizable by a deterministic top-down automaton is a strict subset of that recognized by the bottom-up counterpart, as shown by \cite{tata2007}. On the other hand, the two automata are proved to be equivalent in their non-deterministic instantiation.

Given the analogy between hidden Markov models and non-deterministic automata, it is then natural to ask whether a top-down hidden tree Markov model is fully equivalent to its bottom-up counterpart. The answer to this question is, indeed, negative due to some fundamentally different assumptions at the level of conditional independence properties associated to the two parsing directions. Such assumptions ensure that the two models have different capabilities in terms of probability distributions that can be represented by the corresponding graphical models. The choice of a top-down direction ensures that the corresponding Markov model is characterized by causal relationships from the parent to each child, independently. A bottom-up process, instead, defines a Markov model characterized by so-called v-structure \citep{lauritzen1996graphical}, where the causal relationship points from each child towards the common parent. This creates differences in the local Markov properties which influences the way the nodes exchange information during inference and learning \citep{lauritzen1996graphical}. For instance, v-structures induce controlled information sharing among siblings which is not available in the top-down model. Further, in the latter approach, a node cannot distinguish between its parent and its children, given that the associated causation relationships are probabilistically equivalent, which makes the top-down model less prone to capture structural information in the learned tree distribution. Because of these differences, particular care must be taken in choosing a top-down or bottom-up model whose bias is coherent with the nature of the tree data whose probability we are trying to model.

The paper provides an overview of the hidden tree Markov models in literature, focusing in particular on the underlying conditional independence assumptions and how these impact their suitability to perform learning on specific tree data. We then discuss how such Markov models can be extended to model tree transductions as conditional distributions over trees. Finally, we introduce a non-parametric generalization of the model in Bayesian sense which can be interpreted as a tree automaton with infinite states.

\section{Hidden Markov Models for Tree Structured Data}
\label{sec:htmm}
Before discussing the details of the generative tree models, we briefly summarize the notation used throughout the paper. We focus on labelled rooted trees, denoted as $\mathbf{x}^{n}$ or $\mathbf{y}^{n}$, where the superscript identifies the $n$-th sample tree (in the dataset), consisting of a set of nodes $\mathcal{U}_{n} = \{1,\dots,U_n\}$. The term $u \in \mathcal{U}_{n}$ is used to denote a generic tree node, whose direct ancestor, called \emph{parent}, is denoted as $pa(u)$. A node $u$ can have a variable number of direct descendants (\emph{children}), such that the $l$-th child of node $u$ is denoted as $ch_l(u)$. We assume trees to have maximum finite out-degree $L$ (i.e. the maximum number of children of a node), although the finiteness assumption can be relaxed in non-parametric models (such as those discussed in Section \ref{sec:bayes}). Each vertex $u$ in the tree is associated with a label $x_u$ which, for the purpose of this paper, is supposed to be chosen from a discrete and finite alphabet.

Modeling of a tree distribution $P(\mathbf{x})$ by means of an {\it Hidden Tree Markov Model} (HTMM) is achieved, as in the sequential case, by postulating the existence of an hidden generative process regulated by unobserved Markov state random variables $Q_u$. These follow the same indexing as the nodes $u$ and assume values over the discrete and finite set $[1,\dots,C]$: Section \ref{sec:bayes} will show how we can obtain a model for $C \rightarrow \infty$. The hidden generative process is realized through a transition distribution on the latent states, whose nature and direction follows from the direction of tree parsing. Observable node labels $x_u$ are generated as in the sequential case through a state-conditional mission distribution.

\subsection{Parsing Direction and Causality Assumptions}
\label{sec:basic}
The direction of the process generating the tree determines the properties and complexity of the HTMM state transition function, influencing the type of structural knowledge that can be captured by the hidden states. This, in turn, affects the tree distributions learnable by the model as well as the tree language that is recognized by the corresponding automaton.

\begin{figure}[t]
\begin{center}
\subfigure[\label{fig:tree}]{
	\scalebox{0.45}{
	\begin{tikzpicture}[every node/.style = {minimum size = 1cm, inner sep=0}]
	\node[visible] (xu) at (0,0) {$x_u$};
	\node[visible] (x1) at (-3,-1.5) {$x_1$};
	\node[visible] (x2) at (-1.5,-1.5) {$x_{ch_{l-1}}$};
	\node[visible] (x3) at (0,-1.5) {$x_{ch_l}$};
	\node[visible] (x4) at (1.5,-1.5) {$x_{ch_{l+1}}$};
	\node[visible] (x5) at (3,-1.5) {$x_{ch_L}$};
	\node at (-2.22,-1.5) {$\dots$};
	\node at (2.28,-1.5) {$\dots$};
	\draw[myLink] (xu) to (x1.north);
	\draw[myLink] (xu) to (x2.north);
	\draw[myLink] (xu) to (x3.north);
	\draw[myLink] (xu) to (x4.north);
	\draw[myLink] (xu) to (x5.north);
	\end{tikzpicture}
	}
}
\subfigure[\label{fig:td}]{
	\scalebox{0.45}{
	\begin{tikzpicture}[every node/.style = {minimum size = 1cm, inner sep=0},]
	\node[hidden] (qu) at (0,0) {$Q_u$};
	\node[hidden] (q1) at (-3,-1.5) {$Q_1$};
	\node[hidden] (q2) at (-1.5,-1.5) {$Q_{ch_{l-1}}$};
	\node[hidden] (q3) at (0,-1.5) {$Q_{ch_l}$};
	\node[hidden] (q4) at (1.5,-1.5) {$Q_{ch_{l+1}}$};
	\node[hidden] (q5) at (3,-1.5) {$Q_{ch_L}$};
	\node[visible] (xu) at (2.2,0) {$x_u$};
	\node[visible] (x1) at (-2.5,-3) {$x_1$};
	\node[visible] (x2) at (-1,-3) {$x_{ch_{l-1}}$};
	\node[visible] (x3) at (0.5,-3) {$x_{ch_l}$};
	\node[visible] (x4) at (2,-3) {$x_{ch_{l+1}}$};
	\node[visible] (x5) at (3.5,-3) {$x_{ch_L}$};
	\node at (-2.22,-1.5) {$\dots$};
	\node at (2.28,-1.5) {$\dots$};
	\draw[myLink] (qu) to (q1.north);
	\draw[myLink] (qu) to (q2.north);
	\draw[myLink] (qu) to (q3.north);
	\draw[myLink] (qu) to (q4.north);
	\draw[myLink] (qu) to (q5.north);
	\draw[myLink] (qu) to (xu);
	\draw[myLink] (q1) to (x1.north);
	\draw[myLink] (q2) to (x2.north);
	\draw[myLink] (q3) to (x3.north);
	\draw[myLink] (q4) to (x4.north);
	\draw[myLink] (q5) to (x5.north);
	\end{tikzpicture}
	}
}
\subfigure[\label{fig:siblings}]{
	\scalebox{0.45}{
	\begin{tikzpicture}[every node/.style = {minimum size = 1cm, inner sep=0},]
	\node[hidden] (qu) at (0,0) {$Q_u$};
	\node[hidden] (q1) at (-3,-1.5) {$Q_1$};
	\node[hidden] (q2) at (-1.5,-1.5) {$Q_{ch_{l-1}}$};
	\node[hidden] (q3) at (0,-1.5) {$Q_{ch_l}$};
	\node[hidden] (q4) at (1.5,-1.5) {$Q_{ch_{l+1}}$};
	\node[hidden] (q5) at (3,-1.5) {$Q_{ch_L}$};
	\node[visible] (xu) at (2.2,0) {$x_u$};
	\node[visible] (x1) at (-2.5,-3) {$x_1$};
	\node[visible] (x2) at (-1,-3) {$x_{ch_{l-1}}$};
	\node[visible] (x3) at (0.5,-3) {$x_{ch_l}$};
	\node[visible] (x4) at (2,-3) {$x_{ch_{l+1}}$};
	\node[visible] (x5) at (3.5,-3) {$x_{ch_L}$};
	%\node at (-2.22,-1.5) {$\dots$};
	%\node at (2.28,-1.5) {$\dots$};
	\draw[myLink] (qu) to (q1.north);
	\draw[myLink,dotted] (q1) to (q2);
	\draw[myLink] (q2) to (q3);
	\draw[myLink] (q3) to (q4);
	\draw[myLink,dotted] (q4) to (q5);
	\draw[myLink] (qu) to (xu);
	\draw[myLink] (q1) to (x1.north);
	\draw[myLink] (q2) to (x2.north);
	\draw[myLink] (q3) to (x3.north);
	\draw[myLink] (q4) to (x4.north);
	\draw[myLink] (q5) to (x5.north);
	\end{tikzpicture}
	}
}
\subfigure[\label{fig:bu}]{
	\scalebox{0.45}{
	\begin{tikzpicture}[every node/.style = {minimum size = 1cm, inner sep=0},]
	\node[hidden] (qu) at (0,0) {$Q_u$};
	\node[hidden] (q1) at (-3,-1.5) {$Q_1$};
	\node[hidden] (q2) at (-1.5,-1.5) {$Q_{ch_{l-1}}$};
	\node[hidden] (q3) at (0,-1.5) {$Q_{ch_l}$};
	\node[hidden] (q4) at (1.5,-1.5) {$Q_{ch_{l+1}}$};
	\node[hidden] (q5) at (3,-1.5) {$Q_{ch_L}$};
	\node[visible] (xu) at (2.2,0) {$x_u$};
	\node[visible] (x1) at (-2.5,-3) {$x_1$};
	\node[visible] (x2) at (-1,-3) {$x_{ch_{l-1}}$};
	\node[visible] (x3) at (0.5,-3) {$x_{ch_l}$};
	\node[visible] (x4) at (2,-3) {$x_{ch_{l+1}}$};
	\node[visible] (x5) at (3.5,-3) {$x_{ch_L}$};
	\node at (-2.22,-1.5) {$\dots$};
	\node at (2.28,-1.5) {$\dots$};
	\draw[myLink] (q1.north) to (qu);
	\draw[myLink] (q2.north) to (qu);
	\draw[myLink] (q3.north) to (qu);
	\draw[myLink] (q4.north) to (qu);
	\draw[myLink] (q5.north) to (qu);
	\draw[myLink] (qu) to (xu);
	\draw[myLink] (q1) to (x1.north);
	\draw[myLink] (q2) to (x2.north);
	\draw[myLink] (q3) to (x3.north);
	\draw[myLink] (q4) to (x4.north);
	\draw[myLink] (q5) to (x5.north);
	\draw[myLink2,bend left=27, dotted] (q1) to (q5);
	\draw[myLink2,bend right, dotted] (q1) to (q4);
	\draw[myLink2,bend right, dotted] (q1) to (q3);
	\draw[myLink2,bend right, dotted] (q2) to (q5);
	\draw[myLink2,bend right, dotted] (q2) to (q4);
	\draw[myLink2,bend right, dotted] (q3) to (q5);
	\draw[myLink2,bend left, dotted] (q1) to (q2);
	\draw[myLink2,bend left, dotted] (q2) to (q3);
	\draw[myLink2,bend left, dotted] (q3) to (q4);
	\draw[myLink2,bend left, dotted] (q4) to (q5);
	\end{tikzpicture}
	}
}
\caption{Graphical representation (a) for a labeled subtree rooted in $y_u$, with $L$ children labeled as $y_{ch_l}$. Corresponding DBNs for a top-down HTMM (b), a sibling-dependent HTMM (c) and a bottom-up HTMM (d): empty circles denote hidden variables, while shaded nodes identify observations.}
\label{fig:treeModels}
\end{center}
\vspace{-6mm}
\end{figure}
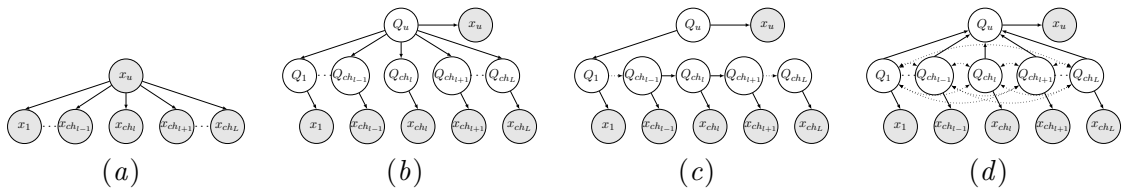

The HTMM model has been introduced almost coincidentally both in its top-down (TD) \citep{hmt98} and bottom-up (BU) \citep{frasconi1998general} form, although the latter has been only sketched, at least initially, without any actual development or application due to computational complexity associated with the transition function. The TD model by \cite{htmm02}, and its variant by \cite{durand2004computational}, have been for several years the sole HTMM approach used in practice. The TD HTMM implements a generative process for all paths from the root to leaves of the trees, modeled by the multinomial \emph{state transition probability} $P(Q_{u} = j | Q_{pa(u)} = i)$ and a prior (multinomial) distribution on the state of the root $P(Q_{1} = j)$. To complete the specification of the model, it is assumed that the node label $x_u$ is completely specified by its hidden state $Q_u$ through the \emph{emission distribution} $P(x_u = m | Q_u = j)$.  \figureref{fig:td} depicts the Dynamic Bayesian Network (DBN) showing the parent-to-children causal dependencies for a TD HTMM corresponding to the exemplar subtree in \figureref{fig:tree}. Following the graphical models convention (see \cite{lauritzen1996graphical}), nodes denote random variables either latent (empty circles) or observable (shaded); an arrow connecting two nodes denotes a conditional dependence of the variable in the destination node given the one associated to the source node. By following the independence relations in the DBN it is possible to factorize the joint TD distribution of an observed tree $\mathbf{x}^n$ as
\begin{equation}\label{eq:THTMM}
    P (\mathbf{x}^n) =  \sum_{i_1, \dots, i_{U_n}} P(Q_{1} = i_1) P(x_{1} | Q_{1} = i_1) \prod_{u = 2}^{U_n} P(x_u | Q_u = i_u)  P(Q_u = i_u | Q_{pa(u)} = i_{pa(u)} ).
\end{equation}
The conditional independence assumptions described by \figureref{fig:td} and \equationref{eq:THTMM} entail that sibling nodes (i.e. those with a common parent) are mutually independent when the parent state is observed. This ensure that TD can capture indirect dependencies between a node and his descendants in the tree by memorizing them into its state transitions. On the other hand, such relationships also state that sibling nodes are conditionally independent given the parent. As a result, TD cannot model any form of dependency between siblings. \cite{ordtree06} have made an attempt to capture such relationship by introducing a causal dependency between neighboring siblings. As shown in \figureref{fig:siblings} this produces a DBN where each non-root node is conditionally dependent either on its ancestor or on its preceding sibling. Such an approach can capture indirect dependencies between sibling nodes but its application is restricted to \emph{ordered trees}, which, are less expressive than the positional trees typically used in practical applications. 

The development of BU generative models for tree data has been limited for many year by the computational issues associated with a state-transition function modeled through the joint probability $P(Q_u = i | Q_{ch_1(u)} = j_1, \dots, Q_{ch_L(u)} = j_L)$, which assumes that each node $u$ is conditionally independent of the rest of the tree when the joint hidden state of its direct descendants $Q_{ch_l(u)} = j_l$ is observed. In other words, each node is evaluated in the context of its child subtrees in accordance with a recursive parsing of the structure starting at the leaves and converging at the root. The leaves are at the basis of such recursive formulation and, as such, are associated with an empty context, modeled by the prior distribution $P(Q_u = i)$.  \figureref{fig:bu}
shows the DBN associated with the BU model, highlighting information sharing among siblings as dashed arrows (resulting from the moralization of the DBN as explained in \cite{lauritzen1996graphical}). In fact, the presence of a v-structure in the DBN ensures that the child nodes $Q_{ch_l}$ are mutually dependent when $Q_u$ is observed, as knowledge regarding the realization of a dependent variable (i.e. $Q_u$) \emph{explains away}, as described by \cite{lauritzen1996graphical}, information about its causes (i.e. the state of the children $Q_{ch_l}$). Therefore, the state transitions of the BHTMM model are capable of capturing indirect dependencies both between a node and its descendants, as well as between sibling subtrees, without losing representational power. 

The direction of conditional dependence influences the capability of a model in capturing structural ancestor-descendant relationships. From a probabilistic point of view, the causation relationship $Q_{pa(u)} \rightarrow Q_{u}  \rightarrow Q_{ch_l(u)}$, which characterizes TD models, is indistinguishable from 
$Q_{pa(u)} \leftarrow Q_{u}  \rightarrow Q_{ch_l(u)}$. This entails that a node in TD cannot probabilistically distinguish between a child and its parent, with clear consequences on the capability of capturing structural information. The BU model, instead, is characterized by a v-structure where conditional dependencies point from all children to the parent. This conditional relationship is probabilistically distinguishable from the bottom-up chain $Q_{pa(u)} \leftarrow Q_{u}  \leftarrow Q_{ch_l(u)}$, hence BU nodes can differentiate between ancestors and descendants.

Recently, \cite{BacciuMS12} have addressed the computational issues associated with the BU HTMM model by exploiting a mixture of multinomials approximation of the joint state transition function, yielding the following factorization of the likelihood of a tree $\mathbf{x}^n$
\begin{equation}\label{eq:BU}
\begin{split}
    P (\mathbf{x}^n) & =  \sum_{i_1, \dots, i_{U_n}} \prod_{u' \in \mathcal{LF}_n} P(Q_{u'} = i_{u'}) P(x_{u'} | Q_{u'} = i_{u'}) \\
    & \times \prod_{u \in  \mathcal{U}_n \setminus \mathcal{LF}_n} P(x_u | Q_u = i_u) \sum_{l = 1}^{L} P(S_u = l) P(Q_u = i_u | Q_{ch_{l}(u)} = i_{ch_l(u)}).
\end{split}
\end{equation}
\equationref{eq:BU} states that the joint children-to-parent transition can be approximated as a mixture of $L$ elementary distributions $P(Q_u = i | Q_{ch_{l}(u)} = j_l)$ where the influence of the $l$-th children on state transition to node $u$ is determined by the weight $P(S_u = l)$. This produces a reduction in the complexity (and number of parameters) of the transition distribution from $O(C^L)$ to $O(LC^2 + L)$, paving the way to a number of advanced BU models, such as generative topographic mappings for tree data \citep{BacciuMS13} and the recent mixture of hidden trees \citep{esann2018Tree}.

Learning in the HTMM models is performed by an iterative maximization process of the likelihood (e.g. \equationref{eq:THTMM,eq:BU}) based on the Expectation Maximization algorithm. Training is based on an alternate optimization estimating the posteriors of the hidden state variables, computed efficiently by recursive upwards-downwards message passing on the structure of the tree, and reusing these to update the current estimate of the distribution parameters.

\subsection{Learning Isomorph Transductions}
\label{sec:io}
In the previous section, we have discussed how HTMMs can be interpreted both as generative models learning unconditional tree distributions $P(\mathbf{x})$, as well as nondeterministic finite state PA for tree languages.  \cite{bengio1996algorithm} have shown that by introducing input-conditional state and emission distribution in HMM these can be extended to learn a probabilistic transducer for sequences. \cite{bacciuNC13} have extended the HTMM along the same lines, showing how their Input-Output HTMM (IO-HTMM) can be used to learn a probabilistic transducer for trees. In particular, the IO-HTMM model is shown to learn a conditional probability $P(\mathbf{y}^n | \mathbf{x}^n)$ that captures the structured transduction from the input trees $\mathbf{x}^n$ to the target structures $\mathbf{y}^n$. A general tree transduction \citep{frasconi1998general} $\tau : \mathcal{X} \rightarrow \mathcal{Y}$ is a binary relation transforming an input tree $\mathbf{x} \in \mathcal{X}$ into an output tree $\mathbf{y}=\tau(\mathbf{x}) \in \mathcal{Y}$. IO-HTMM can be used to learn a restricted class of such relations, including isomorph transductions (i.e. where the output structure has the same topology as the input on) and simple forms of non-isomorph transductions.

Again, when moving from the sequential to the tree-structured domain, the direction of the parsing process becomes crucial in determining the capabilities of the resulting transducer. \cite{Engelfriet75} has shown that the top-down and bottom-up non-deterministic tree transducers have well defined differences in their expressive power. In particular, a top-down transducer cannot accomplish a transformation that creates multiple copies of the path being followed while ensuring that they remain identical until the leaves (recall that a top-down follows all paths independently). Bottom-up can implement such transduction and also the ``checking followed by deletion'', which is also not accomplishable by a top-down tree transducer. Following such considerations, the BU approach seems the most promising to realize a generative model for learning unrestricted tree transductions (which is still an open problem in literature).

An interesting application of the isomorph transductive process has been proposed by \cite{bidir}, which introduced a tree autoencoder model consisting of an encoding module, realized by means of a TD HTMM, and a decoding module realized through an IO-HTMM based on a BU generative process. The fusion of two generative directions (TD and BU) in the autoencoder allows to learn an unconditional tree distribution $P(\mathbf{x})$ which captures richer structural information with respect to an unidirectional BU or TD distribution.

\section{Learning Tree Automata with Infinite States}
\label{sec:bayes}
Finite Markov models are parameterized by the term $C$, which determines the number of different hidden states a random variable $Q_u$ can assume. This value is an hyper-parameter of the learning procedure and therefore should be determined a priori. In general, it is not easy to identify optimal values of model hyper-parameters and we would like the model to be able to adaptively determine the most suitable size of the hidden space. To this end, we can allow the model to have a countably infinite number of hidden states (i.e. $C \rightarrow \infty$). These form of Bayesian models are referred to as nonparametric. In particular, we are interesting in a Bayesian nonparametric (BNP) extension, which provides a Bayesian framework to adapt the nonparametric model to the data \citep{orbanz2011bayesian}. 

\cite{beal2002infinite} introduced a BNP extension of HMM on sequences (iHMM); later, \cite{teh2006hierarchical} related the iHMM to the Hierarchical Dirichlet Process (HDP). Due to the stick-breaking construction of HDP, only a finite number of hidden state are used: the model is able to create new active states on the fly if they are needed to describe the observed data. Given the equivalence stated in \cite{dupont} between PNFA and finite HMM, we can argue that the iHMM is equivalent to the a PNFA with a countably infinite number of states. \cite{pfau2010probabilistic} proposed a Bayesian nonparametric approach to learn a probabilistic deterministic automaton with an infinite number of states (PDIA). Furthermore, they showed that a PDIA is strictly more expressive than a PDFA, but strictly less expressive than a PNFA (and then also HMM).

The Bayesian nonparametric extension are not limited to model sequences. \cite{liang2007infinite} introduced an infinite Probabilistic Context Free Grammar (iPCFG): the number of grammar symbols is not fixed, but it is learned from data. The model defines a distribution over binary trees, whose induced language expresses the derivation trees language of a Chomsky Normal Form PCFG \citep{tata2007}. However, this model is slightly different from a TD HTMM since it allows hidden states with no associated visible label. \cite{finkel2007infinite} introduced three different version of infinite TD HTMM: all of them are obtained using the HDP theory, but assuming a different independence assumptions among children. The first one generates the states of all of the children of a node $u$ (no independence assumption); the second one assume a first-order process to generate the children (Markovian independence assumption); the third one generates children independently of each other (conditional independence assumption). The latter one is the infinite extension of TD HTMM defined in the previous section. Unfortunately, there is no BNP extension for the BU HTMM in literature: we introduce its formalization in the next section.

Again, it is natural to ask the relation between the infinite HTMM and tree automata. We should observe that the differences highlighted in the previous sections are still valid in BNP setting. In fact, the BNP extension only allows the model to have an infinite number of hidden states, but does not affect its independence assumptions. Therefore, we can argue that (i) iTD HTMM is less expressive than infinite TD tree automata (due to the independent child assumption), and (ii) that iBU HTMM represents a BU tree automata with infinite states.

\subsection{Infinite BU HMTMM}
The BNP extension of the BU HMTMM relies on the same theory used in the BNP models discussed before. The key idea is to observe that the transition dynamics can be formalized through a mixture model, where the state of the hidden child determines the mixture component. The iBU HTMM is obtained increasing the number of mixture components to infinity: the result is a HDP.

A HDP is a recursive process based on Dirichlet Process (DP). At the first level of the hierarchy, we sample a global random probability measure $G_0$ distributes as a DP with concentration parameter $\gamma$ and base probability measure $H$. The elements in the second level of the hierarchy are obtained from a DP whose base distribution is $G_0$. The process can be recursively applied in order to obtain deeper hierarchies \citep{teh2006hierarchical}.

We can now introduce the iBU HTMM, using the stick breaking construction of HDP. In the first level we have a global measure $\beta$ which, roughly speaking, indicates the base distribution for the transition dynamics. At the second level we have $L$ different distributions $\beta_l$, one for each position $l$ in the child subtree, allowing the model to have different dynamics for different node positions. The $\beta_l$ are obtained sampling a DP whose base distribution is $\beta$. Finally, we can sample the transition parameters $\pi_{j,l}$ from a DP base on $\beta_l$. The parameters of the emission distribution $\sigma_j$ are sampled from a base distribution $H$: we should sample different parameters for each hidden state $j$. Also, we should provide a priori for SP variables: since they are distributed as a multinomial, the priori should be Dirichlet on the $L$-simplex. The iBU HTMM model can be summarized  by the following schema:
\begin{center}
	\begin{minipage} [t]{0.49\textwidth}
		\begin{align*}
			\beta \mid \gamma &\sim GEM(\gamma)\\
			\beta_l \mid \alpha_l, \beta &\sim DP(\alpha_l, \beta)\\
			\pi_{j,l} \mid \alpha_t, \beta_l &\sim DP(\alpha_t,\beta_l)\\
			\sigma_j &\sim H
		\end{align*}
	\end{minipage}
	\hfill
	\begin{minipage} [t]{0.49\textwidth}
		\begin{align*}
			\phi &\sim Dir(\alpha_{s})\\
			Q_u \mid Q_{ch_l(u)} = j, \pi_{j,l} &\sim \text{Mult}(\pi_{j,l})\\
			y_u \mid Q_u = j, \sigma_j  &\sim \text{Mult}(\sigma_j)
		\end{align*}
	\end{minipage}
	
	\[
		Q_u \mid Q_{ch_1(u)} = j_1, \pi_{j_1,1}, \dots, Q_{ch_L(u)} = j_L, \pi_{j_L,L} \sim \sum_{l=1}^L P(S_u=l\mid\phi) \times \pi_{j_l,l}.
	\]
\end{center}

\cite{teh2006hierarchical} proposed three different inference procedure to learn the parameters of HDP, which can be extended to learn iHMM both on sequences and trees. \cite{finkel2007infinite} used the direct assignment defined in \cite{teh2006hierarchical} to learn the parameters of infinite trees.
Nevertheless, \cite{van2008beam} and \cite{tripuraneni2015particle} proposed new methods to sample the hidden trajectory in iHMM for sequences. In particular, the method introduced by \cite{tripuraneni2015particle} seems to be more suitable to do inference on tree data structures.
\begin{figure}[t]
	\centering
	\subfigure[\label{fig:inf-TD-HTMM}Infinite TD HMM \citep{finkel2007infinite}]
	{%						
		\scalebox{0.7}
		{%		
			\begin{tikzpicture}[every node/.style = {circle,  minimum size = 1cm}]
			\begin{scope}
			\node[draw] (g) at (0,2) {$\gamma$};
			\node[draw] (b) at (2,2) {$\beta$};
			\node[draw] (a0) at (0,0) {$\alpha_0$};
			\node[draw] (pi) at (2,0) {$\pi_jl$};
			\node[draw] (h) at (0,-2) {$H$};
			\node[draw] (s) at (2,-2) {$\sigma_j$};
			\draw  (1,1) rectangle (3,-3);
			\node at (2.8,-2.8) {$\infty$};
			\draw[myLink] (g) to (b);
			\draw[myLink] (g) to (b);
			\draw[myLink] (a0) to (pi);
			\draw[myLink] (b) to (pi);
			\draw[myLink] (h) to (s);
			\end{scope}
			\begin{scope}[shift={(-2,1.6)}]
			\node[draw] (q1) at (6,0) {$Q_1$};
			\node[draw] (q2) at (7.6,-1.6) {$Q_2$};
			\node[draw] (q3) at (9,-1.6) {$Q_3$};
			\node[visible] (x1) at (6,-3.6) {$x_1$};
			\node[visible] (x2) at (7.6,-3.6) {$x_2$};
			\node[visible] (x3) at (9,-3.6) {$x_3$};
			\draw[myLink] (q1) to (q2);
			\draw[myLink] (q1) to (q3);
			\draw[myLink] (q1) to (x1);
			\draw[myLink] (q2) to (x2);
			\draw[myLink] (q3) to (x3);
			\draw[myLink,dotted, bend left] (s) to (x1);
			\draw[myLink,dotted, bend left] (s) to (x2);
			\draw[myLink,dotted, bend left] (s) to (x3);
			\draw[myLink,dotted, bend left] (pi) to (q2);
			\draw[myLink,dotted, bend left] (pi) to (q3);
			\end{scope}
			\end{tikzpicture}		
		}
	}
	\hfill
	\subfigure[\label{fig:inf-BU-HTMM}Infinite BU HTMM]
	{%						
		\scalebox{0.7}
		{%						
			\begin{tikzpicture}[every node/.style = {circle,  minimum size = 1cm}]			
			\begin{scope}
			\node[draw] (g) at (0,2) {$\gamma$};
			\node[draw] (b) at (2,2) {$\beta$};
			\node[draw] (al) at (0,0) {$\alpha_l$};
			\node[draw] (bl) at (2,0) {$\beta_l$};
			\node[draw] (atr) at (0,-2) {$\alpha_t$};
			\node[draw] (pi) at (2,-2) {$\pi_{j,l}$};
			\node[draw] (s) at (2,-3.6) {$\sigma_j$};
			\node[draw] (h) at (4,-3.6) {$H$};
			\node[draw] (as) at (4,2) {$\alpha_s$};
			\node[draw] (phi) at (4,0) {$\phi$};
			\draw  (1,1) rectangle (3,-3);
			\node at (2.8,0.8) {$L$};
			\draw  (1.2,-1) rectangle (2.8,-4.6);
			\node at (1.4,-4.4) {$\infty$};
			\draw[myLink] (g) to (b);
			\draw[myLink] (g) to (b);
			\draw[myLink] (al) to (bl);
			\draw[myLink] (b) to (bl);
			\draw[myLink] (atr) to (pi);
			\draw[myLink] (bl) to (pi);
			\draw[myLink] (h) to (s);
			\draw[myLink] (as) to (phi);
			\end{scope}	
			\begin{scope}
			\node[draw] (q1) at (6,0) {$Q_1$};
			\node[draw] (q2) at (7.6,-1.6) {$Q_2$};
			\node[draw] (q3) at (9,-1.6) {$Q_3$};			
			\node[visible] (x1) at (6,-3.6) {$x_1$};
			\node[visible] (x2) at (7.6,-3.6) {$x_2$};
			\node[visible] (x3) at (9,-3.6) {$x_3$};		
			\draw[myLink] (q2) to (q1);
			\draw[myLink] (q3) to (q1);
			\draw[myLink] (q1) to (x1);
			\draw[myLink] (q2) to (x2);
			\draw[myLink] (q3) to (x3);		
			\draw[myLink,dotted, bend left] (s) to (x1);
			\draw[myLink,dotted, bend left] (s) to (x2);
			\draw[myLink,dotted, bend left] (s) to (x3);			
			\draw[myLink,dotted, bend right] (pi) to (q1);
			\draw[myLink,dotted, bend left] (phi) to (q1);
			\end{scope}
			\end{tikzpicture}	
		}
	}
\label{fig:BNP-tree}
\caption{A graphical representation of the infinite TD HTMM and BU HTMM.}
\end{figure}
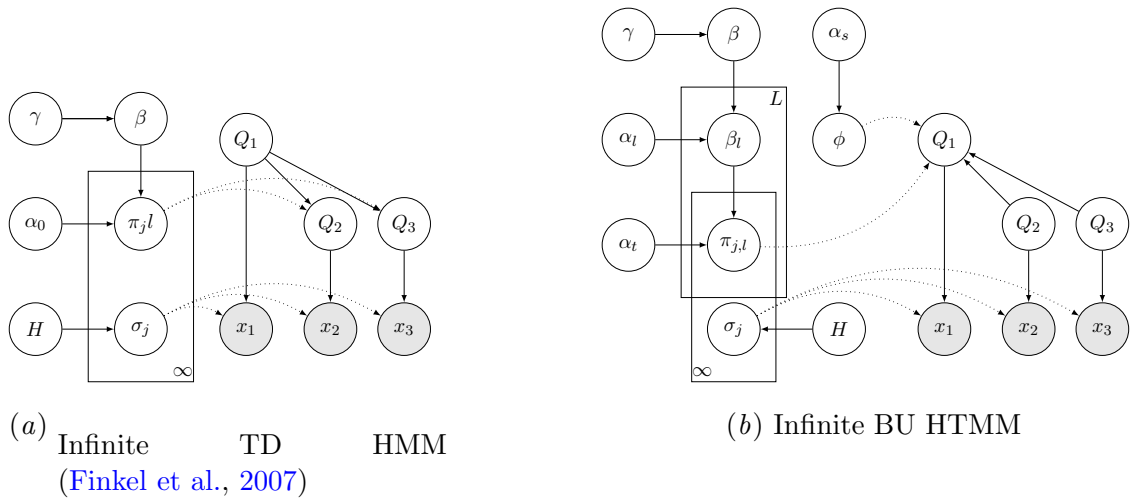

\section{Conclusions}
\label{sec:conc}

The paper provides an overview of the Hidden Tree Markov Models in literature. 
In particular, we show how the parallelism between HMM and NDFA can lead to erroneous conclusion when we are dealing tree data structures. The different independence assumptions associated to TD HTMM and BU HTMM makes the two generative processes not equivalent, while the nondeterministic TD tree automata is proven to be equivalent to nondeterministic BU tree automata.

The ability of share information among siblings, that is an inherent property of the independence assumptions embodied in the BU generative process, allows the BU HTMM to better capture structural information. We argue this could be a starting point to create a generative model for leaning unrestricted tree transductions. In the last section of the paper, we have sketched a novel BNP extension of the BU HTMM, showing how it can be used to learn the hidden state number, which is an hyperparameter of BU HTMM, directly from data. We have discussed how this model can be interpreted as a BU HTMM with infinite states, leading to a potentially interesting interpretation of this model as nondeterministic PA for tree language with infinite states.

\acks{The work is partially funded by Italian Ministry of Education, University, and Research (MIUR) under project SIR 2014 LIST-IT (grant n. RBSI14STDE)}

\bibliography{markovtree}

\end{document}